\documentclass[runningheads,a4paper]{llncs}
\usepackage[bookmarks=true]{hyperref}
\usepackage{amssymb}
\usepackage{marvosym}
\usepackage{graphicx}
\usepackage{array}
\usepackage{amsmath}
\usepackage[table]{xcolor}
\usepackage{graphicx}
\usepackage{ifthen}
\usepackage{changepage}
\usepackage{multicol}
\usepackage{multirow}
\usepackage{pgfplots}
\usepackage{algorithm}
\usepackage{url}
\definecolor{entete}{RGB}{221,221,221}
\definecolor{col}{RGB}{238,238,238}
\definecolor{Ver}{RGB}{29,71,109}
\definecolor{Hig}{RGB}{48,120,178}
\definecolor{Mo}{RGB}{152,193,237}
\definecolor{Lo}{RGB}{131,150,183}
\definecolor{vLo}{RGB}{196,204,217}
\definecolor{cVer}{RGB}{79,124,55}
\definecolor{cHig}{RGB}{168,208,138}
\definecolor{cMo}{RGB}{199,224,182}
\definecolor{cLo}{RGB}{191,143,0}
\definecolor{cvLo}{RGB}{128,95,0}
\usepackage{soul}

\usepackage{epstopdf}	
\usepackage[noend]{algpseudocode}
\usepackage{tikz,pgfplots}
\usepackage{caption}
\usetikzlibrary{calc,positioning,shapes.geometric,backgrounds,fit}
\urldef{\mailsa}\path|{alfred.hofmann, ursula.barth, ingrid.haas, frank.holzwarth,|
\urldef{\mailsb}\path|anna.kramer, leonie.kunz, christine.reiss, nicole.sator,|
\urldef{\mailsc}\path|erika.siebert-cole, peter.strasser, lncs}@springer.com|    
\newcommand{\keywords}[1]{\par\addvspace\baselineskip
\noindent\keywordname\enspace\ignorespaces#1}

\begin{document}

\mainmatter  

\title{Expert Opinion Extraction from a Biomedical Database}
\titlerunning{Expert Opinion Extraction from a Biomedical Database}

\author{Ahmed Samet\inst{1} \and Thomas Guyet\inst{1} \and Benjamin Negrevergne\inst{3} \and Tien-Tuan Dao\inst{2} \and Tuan Nha Hoang\inst{2} \and Marie Christine Ho Ba Tho\inst{2}}

\institute{Universit\'{e} Rennes 1/IRISA-UMR6074\\      \email{firstname.lastname@irisa.fr}
\and
Sorbonne University, Universit\'{e} de technologie de Compi\`{e}gne\\ CNRS, UMR 7338 Biomechanics and Bioengineering, Compi\`{e}gne, France \email{\{firstname.lastname\}@utc.fr}
\and
LAMSADE, Université\'{e} Paris-Dauphine\\    \email{firstname.lastname@dauphine.fr}}

\authorrunning{Ahmed Samet et al.}

\maketitle

\begin{abstract}
In this paper, we tackle the problem of extracting frequent opinions from uncertain databases. We introduce the foundation of an opinion mining approach with the definition of pattern and support measure. The support measure is derived from the commitment definition. A new algorithm called OpMiner that extracts the set of frequent opinions modelled as a mass functions is detailed. Finally, we apply our approach on a real-world biomedical database that stores  opinions of experts to evaluate the reliability level of biomedical data. Performance analysis showed a better quality patterns for our proposed model in comparison with literature-based methods.
\end{abstract}

\section{Introduction}

Data uncertainty has challenged nearly all types of data mining tasks, creating a need for uncertain data mining.
Uncertainty is inherent in data from many different domains, including social networks and cheminformatics~\cite{moi15}. The problem of pattern mining, or finding frequent patterns in data, has been extensively studied in deterministic databases~\cite{Agg14} since its introduction by Aggrawal et al. ~\cite{Agr94} as well as in the field of uncertain databases~\cite{Agg09}. The uncertain databases have brought more flexibility in data representation~\cite{Bell96}. For instance, mass function of evidence theory are comparable to expert's opinion since it details answer to a question over a set of response elements. It also allows to model someone's degree of belief regarding an asked question. Therefore databases storing mass functions (commonly called evidential databases), are seen as a data support for expert opinions and imperfect data. 

What classical approaches have in common is that they extract answers. They extract answer elements (fragment of the expert answer) as long they are redundant in the database. Therefore, the extracted information is limited and does not describe what experts have expressed. To illustrate this point, let us consider the example of several practitioners that have been asked to give their opinion regarding new treatments for a disease. We intend to extract knowledge from a set of experts' opinions asked about their evaluations of these treatments. Each practitioner gives his opinion regarding the efficiency of a treatment $j$ among a set of evaluation possibilities $\{Good_j,Average_j,Bad_j\}$.,

\begin{table}
\centering
\begin{tabular}{lcc}
\hline
Practitioner & Treatment 1 & Treatment 2\\
\hline
$P_1$& $Bad_1^{0.3}\ Average_1^{0.7} $&$Good_2^1$\\
$P_2$& $\{Average_1\cup Bad_1\}^{1}$ &$Good_2^{0.5} \ Average_2^{0.5}$\\
\hline
\end{tabular}
\caption{Example of uncertain database.}
\label{example motivation}
\end{table}

The first practitioner hesitates between bad and average evaluation with a higher confidence to average. The second practitioner can not decide whether the treatment is average or bad. A classical pattern mining approach as~\cite{Chui07} would extract answers as pattern. For instance, for a threshold of 0.7, \{$Treatment1=Average_1$\} is a frequent pattern\footnote{A pattern is called frequent if its computed support (i.e. frequency in the database) is higher than or equal to a fixed threshold set by an expert}. Looking at Table \ref{example motivation}, the pattern \{$Treatment1=Average_1$\} is a fraction of the opinion expressed by the practitioner $P_1$ and therefore the extracted information is not complete. Unfortunately, this type of output is generated by uncertain mining approaches~\cite{Agg10,Hew07,Chen08}. An opinion pattern would be $Treatment_1=Bad^{0.3}\ Average^{0.7}$ and is considered as frequent since it does not contradict with the opinion of $P_2$. In this work, we intend to shake this notion of answer pattern of uncertain databases and we aim to evaluate a pattern as a whole opinion.

Methodologically, we build the foundation of an opinion mining approach. We develop our mining approach on evidential databases. Evidential databases offer more knowledge representation with its simple formalism~\cite{Bach09}. They bring more flexibility thanks to mass functions. In fact, it is possible to model all level of uncertainty from absolute certainty to total ignorance. From applicative point of view, we experiment our OpMiner algorithm on a real-world biomedical expert database. The results show the quality of retrieved patterns comparatively to classical ones. In addition, our algorithm shows interesting performances.

\section{Preliminaries}\label{section 1}

The evidence theory or Dempster-Shafer theory~\cite{dem67,Sh76} proposes a robust formalism  for modeling uncertainty. The evidence theory is based on several fundamentals such as the Basic Belief Assignment (BBA). A BBA $m$ is the mapping from elements of the power set $2^\Theta$ onto [0, 1]:
	\[m:2^{\Theta}\longrightarrow [0,1]
\] 
	
where $\Theta$ is the \textit{frame of discernment}. It is the set of possible answers for a addressed problem and is composed of $N$ exhaustive and exclusive hypotheses:
\[
\Theta=\{H_{1},H_{2},...,H_{N}\}.
\]

A BBA $m$ is constrained by:

\begin{equation}
\begin{cases}
\sum_{A\subseteq \Theta}m(A)=1 \\
m(\emptyset)=0 \end{cases}.
\end{equation}

Each subset X of $2^\Theta$ fulfilling $m(X)>0$ is called focal element. Constraining $m(\emptyset)=0$ is the normalized form of a BBA and this corresponds to a closed-world assumption, while allowing $m(\emptyset)>0$ corresponds to an open world assumption\cite{Sm94}.

Dubois and Prade \cite{Dub86b} have made three proposals to order BBAs. Let $m_1$ and $m_2$ be two BBA's on $\Theta$. The statement that $m_1$ is at least as committed as $m_2$ is denoted $m_1 \sqsubseteq m_2$. Three types of ordering have been proposed:
\begin{itemize}
	\item \textit{pl-ordering} (plausibility ordering) if $Pl_1(A)\leq Pl_2(A)$ for all $A \subseteq \Theta$, we write $m_1 \sqsubseteq_{pl} m_2$,
	\item \textit{q-ordering} (communality ordering) if $q_1(A)\leq q_2(A)$ for all $A \subseteq \Theta$, we write $m_1 \sqsubseteq_{q} m_2$,
	\item \textit{s-ordering} (specialization ordering) if $m_1$ is a specialization of $m_2$, we write $m_1 \sqsubseteq_{s} m_2$,
\end{itemize}
In this paper, we develop our approach using the plausibility based commitment. The plausibility function $Pl(.)$ is defined as follows:

\begin{equation}
Pl(A)=\sum_{B\cap A\neq \emptyset}m(B).
 \end{equation}
Among all belief functions on $\Theta$, the least committed belief function is the vacuous belief function (i.e. $m(\Theta)=1$).

Finally, it is possible to store imperfect data modelled as BBAs into a database. This kind of database is commonly called evidential database. Formally, an \textit{evidential database} is a triplet $\mathcal{EDB}=(\mathcal{A_{EDB}},\mathcal{O},R_{\mathcal{EDB}})$.  $\mathcal{A_{EDB}}$ is a set of attributes and $\mathcal{O}$ is a set of $d$ transactions (i.e., rows). Each column $A_j$ $(1 \leq j \leq n)$ has a domain $\Theta_{j}$ of discrete values. $R_{\mathcal{EDB}}$ expresses the relationship between the $i^{th}$ transaction (i.e., row $T_i$) and the  $j^{th}$ column (i.e., attribute $A_j$) by a normalized BBA $m_{ij}:2^{\Theta_{j}} \rightarrow [0,1] $.

\begin{example}\label{example1}
We intend to extract knowledge from a set of experts' opinions asked about their evaluations of several treatment efficiencies for a disease. Each practitioner gives his opinion regarding a treatment $j$ from a set of evaluation possibilities $\Theta_j=\{Good_j,Average_j,Bad_j\}$.

\begin{table}
\centering
\begin{tabular}{lcc}
\hline
Practitioner & Treatment 1  & Treatment 2 \\
\hline
$P_1$ & $m_{11}(Good_1)=0.7$ & $m_{12}(Good_2)=0.4$\tabularnewline
 & $m_{11}(\Theta_1)=0.3$ & $m_{12}(Average_2)=0.2$\tabularnewline
 &  & $m_{12}(\Theta_2)=0.4$\tabularnewline
$P_2$ & $ m_{21}(Good_1)=0.6$& $m_{22}(Good_2)=0.3$\tabularnewline
 &  $ m_{21}(\Theta_1)=0.4$& $m_{22}(\Theta_2)=0.7$\tabularnewline
\hline
\end{tabular}
\caption{Example of evidential database}
\end{table}
\end{example}

\section{Extraction opinion patterns over evidential databases}
In the following subsection, we study the plausibility based commitment relation between two BBAs in the evidence theory. 
\subsection{Plausibility based commitment measure}
Let us consider two BBAs $m_1$ and $m_2$ such as $m_1 \sqsubseteq_{pl} m_2$. We intend to develop a measure to estimate the commitment level of $m_2$ wrt $m_1$.

\begin{definition}
Given the plausibility functions $Pl_1$ and $Pl_2$ of two BBAs $m_1$ and $m_2$, the plausibility $PL_{12}(.)$ expresses the difference between two plausibility functions and is computed as follows:
\begin{equation}
PL_{12}(A)=Pl_1(A)-Pl_2(A).
\end{equation}
\end{definition}

\begin{definition}
Assuming two BBAs $m_1$ and $m_2$. Assuming that $C(\cdot,\cdot)$ is a commitment measure between two BBAs. It is computed as follows,

\begin{equation}\label{distance}
\begin{array}{ll}
C&:2^{\Theta}\times 2^{\Theta}\mapsto [0,1]\\
 & (m_2,m_1) \rightarrow 
\left\{
\begin{array}{ll}
1-||PL_{21}||=1-\sqrt{\sum\limits_{A \subseteq \Theta} PL_{21}(A)^{2}}& if \  m_1 \sqsubseteq_{pl} m_2\\
0 & Otherwise
\end{array}\right.
\end{array}
\end{equation}
\end{definition}

\begin{property}
Assuming two BBAs $m_1$ and $m_2$ such as $m_2 \sqsubseteq_{pl} m_1$, Equation \ref{distance} verifies the following properties:
\begin{itemize}
	\item $C(m_2,m_1)\geq 0$ \textit{(separation axiom)};
	\item $C(m_2,m_1)=1\Leftrightarrow m_1=m_2$  \textit{(identity of indiscernible)};
	\item $C(m_2,m_1)=C(m_1,m_2)$ \textit{(symmetry)};
	\item $C(m_2,m_3)\leq C(m_2,m_1) +C(m_1,m_3)$ \textit{(triangle inequality)}.
\end{itemize}
\end{property}

\subsection{Mining opinions over evidential databases}
In an evidential database, an \textit{item} corresponds to a BBA. An \textit{itemset} (so called \textit{pattern}) corresponds to a conjunction of several BBAs having different domains  $X=\{m_{ij} \in \mathcal{M}^{\Theta}\}$. We recall that $i$ is the transaction id and $j$ is the attribute id. $\mathcal{M}^{\Theta}$ denotes the set of all BBAs in $\mathcal{EDB}$.

Let us consider an evidential database $\mathcal{EDB}$ and the itemset $X$ made of a set of BBAs. The frequency of appearance of an item $x=m_{i'j}$ in a transaction $T_i$ can be computed as follows:
\begin{equation}
\begin{array}{rcl}
	Sup_{T_i}: \mathcal{M}^{\Theta_j}_i &\rightarrow& [0, 1]\\
	 x &\mapsto& C(x,m_{ij}) \text{ where } m_{ij} \in \mathcal{M}^{\Theta_j}_i.
	\end{array}
\end{equation}

$\mathcal{M}^{\Theta_j}_i$ is the set of BBAs in the row $T_i$ of the attribute $j$.
As illustrated above, the $Sup_{T_i}$ is a measure that computes whether $x$ is in the row $T_i$. Even if the BBA is not in the studied row, we analyse if there is a BBA that generalizes it. Then, the support of an itemset $X$ over the transaction $T_i$ is computed as
\begin{equation}
Sup_{T_i}(X)=\prod\limits_{x \in X} Sup_{T_i}(x).
\end{equation}

Therefore, the support of $m_{ij}$ over the database is computed as,
\begin{eqnarray}
Sup_{\mathcal{EDB}}(X)=\frac{1}{d}\sum\limits^d_{ i=1} Sup_{T_i}(X).\label{support total}
\end{eqnarray}

\begin{property}
Assuming an itemset $X$, the measure of support fulfils the anti-monotony property, i.e.,
\begin{equation}
Sup_{\mathcal{EDB}}(X) \leq  Sup_{\mathcal{EDB}}(X\cup m_{ij}).
\end{equation}
 
\end{property}
\begin{proof}

Assuming an evidential database $\mathcal{EDB}$, let us consider two evidential itemsets $X$ and $X\cup m_{ij}$. We aim at proving this relation $Sup_{\mathcal{EDB}}(X)\leq  Sup_{\mathcal{EDB}}(X\cup m_{ij})$:

\begin{equation*}
\begin{array}{l}
Sup_{T_i}(X\cup m_{ij})=\prod\limits_{m_{i'j'} \in X\cup m_{ij}} Sup_{T_i}(m_{i'j'})\\
Sup_{T_i}(X\cup m_{ij})=\prod\limits_{m_{i'j'} \in X} Sup_{T_i}(m_{i'j'}) \times Sup_{T_i}(m_{ij})\\
Sup_{T_i}(X\cup m_{ij})\leq  Sup_{T_i}(X) \quad \text{since} \quad Sup_{T_i}(m_{ij}) \in [0,1]\quad \text{then} \\
Sup_{\mathcal{EDB}}(X\cup m_{ij})\leq  Sup_{\mathcal{EDB}}(X).
\end{array}
\end{equation*}
\end{proof}

\begin{example}
Assuming the evidential database given in Example \ref{example1}.
For a $minsup=0.7$, the pattern $\{m_{11},m_{12}\}$ have a support of $\frac{C(m_{11},m_{11})\times C(m_{12},m_{12})+C(m_{11},m_{21})\times C(m_{12},m_{22})}{2}=0.765$ and is then considered as frequent. Semantically, having a relatively good opinion on treatment 1 (i.e. $m_{11}$) and hesitant one regarding the treatment 2 (i.e. $m_{12}$) is redundant over $76.5\%$ of asked practitioners. Moreover, comparatively to patterns of an evidential data mining algorithm, our output is more informative. In fact, a classical algorithm would provide the frequent pattern $\{Good_1,Good_2\}$ which contain less details than $\{m_{11},m_{12}\}$.
\end{example}

In this section, we develop a new level-wise algorithm to mine opinions over evidential databases. OpMiner, shown in Algorithm \ref{alg2} generates all BBAs of size one by favouring the most specific ones. Formally, for all $m_{ij}, m_{i'j}\in \mathcal{M}^{\Theta}$, we retain $m_{ij}$ as long as $m_{ij} \sqsubseteq_{pl} m_{i'j}$. Thus, function \textit{candidate\_gen} reduces the set of frequent patterns to the set of the more specific ones. The other less specific BBAs are used to compute the support as described in Equation \ref{support total}. In addition, this selection aims at reducing time computing since candidate generation and support computing depends on the set of items (i.e. pattern with a single BBA). The patterns that have a support lower than the $minsup$ are pruned in line 5. The process stops until no candidate is left.
\begin{algorithm*} [!ht]
\caption{OpMiner algorithm}
   \label{alg2}
	\begin{adjustwidth}{-0cm}{}
   \begin{algorithmic}[1] 
   \footnotesize
     \Require $\mathcal{EDB}, minsup, \mathcal{EDB}_{pl},maxlen$
    \Ensure $\mathcal{EIFF}$
     \State $\mathcal{EIFF},Items \gets \emptyset,size \gets 1$
     \State $Items \leftarrow \Call{candidate\_gen}{\mathcal{EDB},\mathcal{EIFF},    Items}$

     \State \textbf{While} ($candidate \neq \emptyset\ and\ size\leq maxlen $)
  \ForAll{$pat \in candidate$} 
	\If{\Call{Support}{$pat, minsup, \mathcal{EDB}_{pl},   Size\_\mathcal{EDB}$}$\geq minsup$}
 
	\State $\mathcal{EIFF} \gets \mathcal{EIFF}\cup pat$ 
	\EndIf
	\EndFor
 \State $size \gets size+1$
	 
	  \State $candidate \leftarrow \Call{candidate\_gen}{\mathcal{EDB},\mathcal{EIFF}     ,Items}$
 \State \textbf{End While}

  \Function{Support}{$pat$, $minsup$, $\mathcal{EDB}_{pl}$,$d$}
  \State $Sup \gets 0 $
	 
	\For{i=1 to d}
	\ForAll{$pl_{ij} \in \mathcal{M}_i$}
	\State $pl\gets mtopl(pat)\backslash\backslash$ \textit{computes the plausibility out of a BBA}
	\If{$ pl_{ij}\geq pl$}
	
  \State $Sup_{Trans} \gets Sup_{Trans}\times 1-||pl_{ij}- pl|| $
   \EndIf 
	\EndFor
	\State $Sup \gets Sup + Sup_{Trans}$
	\EndFor
	\State \Return $\frac{Sup_I}{d}$
  \EndFunction
	
 \Function{candidate\_gen}{$\mathcal{EDB}$, $\mathcal{EIFF}$, $Items$}
\If{$size(Items)=0$}
\ForAll{$BBA \in \mathcal{EDB}$}
\While{$Items\neq \emptyset\ and\ BBA\not\sqsubseteq_{pl} it$}
\If{$Items= \emptyset$}
\State $Add(BBA,Item)$
\Else
\State $Replace(BBA,it,Item)$
\EndIf
\EndWhile
\EndFor
\State \Return $Items$
  \Else
	\ForAll{$BBA \in \mathcal{EIFF}$}
	\ForAll{$it \in Items$}
	\If {$!same\_attribute(it,BBA)$}
	\State $Cand\gets Cand \cup \{BBA\cup it\}$
	\EndIf
	\EndFor
	\EndFor
	\State \Return $Cand$
	\EndIf
	
\EndFunction
       \end{algorithmic}
			\end{adjustwidth}
 \end{algorithm*}
\section{Experiments: Data reliability assessment using biomedical expert opinion}
The investigation of the effects of muscles morphology and mechanics on motion, and the risks of injury, has been at the core of many studies, sometimes with conflicting results. Often different measurement methods have been used, making comparison of the results and drawing sound conclusions impossible~\cite{Hoa16}. In this section, we aim at studying the opinion of several experts on collected measurement data. To do so, we collected data by a systematic review process of 20 data sources (papers) from reliable search engines (PubMed and ScienceDirect). Data is described over 7 parameters regarding muscle morphology, mechanics and motion analysis. 
Four main questions were asked to experts about measuring technique ($Q_1$), experimental protocol ($Q_2$), number of samples ($Q_3$) and range of values ($Q_4$). An expert opinion database was built from an international panel of 20 contacted experts with different expertise (medical imaging, motion analysis). Five evaluation degrees were possible $\{Very\ high, High, Moderate, Low, Very\ low$\}. Each given degree was associated to a confidence value. In this study, as a first goal, we aim at finding frequent opinions in the database. Frequent opinions show correlation between opinions. The second goal is to evaluate the reliability of sources. To do so, the algorithm selects from the frequent set of patterns those that express a positive opinion regarding the same source.

Table \ref{data} shows a small set of recorded answers from experts. For instance, row 1 details the opinions of expert 1 regarding the source S1 (data measures retrieved from a source). The expert expresses his opinion over 4 questions. The column $Conf_i$ shows the confidence of the expert regarding his given opinion for the question $Q_i$.

\begin{table}
\begin{minipage}{.70\linewidth}

\begin{tabular}{crrrrrrrr}
\rowcolor{entete}Expert&\multicolumn{8}{c|}{S1}\\
\rowcolor{entete}& $Q_1$&$Conf_1$&$Q_2$&$Conf_2$& $Q_3$&$Conf_3$&$Q_4$&$Conf_4$\\
\cellcolor{col}1&\cellcolor{Hig}Hig&\cellcolor{cHig}Hig&\cellcolor{Hig}Hig&\cellcolor{cHig}Hig&\cellcolor{Mo}Mo&\cellcolor{cHig}Hig&\cellcolor{Hig}Hig&\cellcolor{cMo}Mo\\
\cellcolor{col}2&\cellcolor{Hig}Hig&\cellcolor{cVer}Ver&\cellcolor{Mo}Mo&\cellcolor{cVer}Ver&\cellcolor{Hig}Hig&\cellcolor{cVer}Ver&\cellcolor{Mo}Mo&\cellcolor{cVer}Ver\\
\cellcolor{col}3&\cellcolor{Hig}Hig&\cellcolor{cHig}Hig&\cellcolor{Hig}Hig&\cellcolor{cHig}Hig&\cellcolor{Hig}Hig&\cellcolor{cHig}Hig&\cellcolor{Hig}Hig&\cellcolor{cHig}Hig\\
\cellcolor{col}4&\cellcolor{Hig}Hig&\cellcolor{cHig}Hig&\cellcolor{Mo}Mo&\cellcolor{cHig}Hig&\cellcolor{Hig}Hig&\cellcolor{cHig}Hig&\cellcolor{Mo}Mo&\cellcolor{cHig}Hig\\
\cellcolor{col}5&\cellcolor{Lo}Lo&\cellcolor{cVer}Ver&\cellcolor{Lo}Lo&\cellcolor{cVer}Ver&\cellcolor{Mo}Mo&\cellcolor{cVer}Ver&\cellcolor{Mo}Mo&\cellcolor{cVer}Ver\\
\cellcolor{col}6&\cellcolor{Mo}Mo&\cellcolor{cMo}Mo&\cellcolor{Mo}Mo&\cellcolor{cMo}Mo&\cellcolor{Lo}Lo&\cellcolor{cHig}Hig&\cellcolor{Lo}Lo&\cellcolor{cHig}Hig\\
\cellcolor{col}7&\cellcolor{Mo}Mo&\cellcolor{cVer}Ver&\cellcolor{Mo}Mo&\cellcolor{cVer}Ver&\cellcolor{Hig}Hig&\cellcolor{cVer}Ver&\cellcolor{Mo}Mo&\cellcolor{cVer}Ver\\
\cellcolor{col}8&\cellcolor{Mo}Mo&\cellcolor{cVer}Ver&\cellcolor{Lo}Lo&\cellcolor{cHig}Hig&\cellcolor{Hig}Hig&\cellcolor{cVer}Ver&\cellcolor{Lo}Lo&\cellcolor{cVer}Ver\\
\cellcolor{col}9&\cellcolor{Mo}Mo&\cellcolor{cVer}Ver&\cellcolor{Mo}Mo&\cellcolor{cHig}Hig&\cellcolor{Hig}Hig&\cellcolor{cVer}Ver&\cellcolor{Mo}Mo&\cellcolor{cHig}Hig\\
\cellcolor{col}10&\cellcolor{Mo}Mo&\cellcolor{cHig}Hig&\cellcolor{Mo}Mo&\cellcolor{cHig}Hig&\cellcolor{Mo}Mo&\cellcolor{cHig}Hig&\cellcolor{Mo}Mo&\cellcolor{cHig}Hig\\
\cellcolor{col}11&\cellcolor{Ver}Ver&\cellcolor{cVer}Ver&\cellcolor{Ver}Ver&\cellcolor{cVer}Ver&\cellcolor{Ver}Ver&\cellcolor{cVer}Ver&\cellcolor{Ver}Ver&\cellcolor{cVer}Ver\\
\end{tabular}

\end{minipage}
\begin{minipage}{.18\linewidth}
\small
\begin{tabular}{p{1.5cm}p{1.5cm}}
\cellcolor{Ver} Very high& \cellcolor{cVer} Very high confidence\\
\cellcolor{Hig} High& \cellcolor{cHig}  High confidence\\
\cellcolor{Mo} Moderate& \cellcolor{cMo} Moderate confidence\\
\cellcolor{Lo} Low& \cellcolor{cLo}  Low confidence\\
\cellcolor{vLo} Very low& \cellcolor{cvLo}  Very low confidence\\
\end{tabular}

\end{minipage}

\caption{Sample of the expert opinion data.}
\label{data}
\end{table}
	
The evidential database is constructed by using the evaluation of the experts and their confidences. First, the evaluation of the expert is used to model a certain BBA\footnote{A BBA is called a certain BBA when it has one focal element, which is a singleton. It is representative of perfect knowledge and the absolute certainty.}. Then, the confidence is used to integrate uncertainty into the BBA. To do so, the confidence is used as reliability measure and part of the mass initially given to the evaluation is then transferred to the ignorance mass. Formally, the discounting of a mass function $m$ can be written as follows
\begin{equation}
\begin{cases}
m^\alpha(B)=(1-\alpha)\times m(B) & \forall B\subseteq\Theta\\
m^\alpha(\Theta)=(1-\alpha)\times m(\Theta)+\alpha.
\end{cases} 
\label{equ:disc} 
\end{equation}
$\alpha$ is the reliability factor and is in the set $\{0.8,0.6,0.4,0.2,0\}$. The higher $\alpha$ is the more mass is transferred to  $m(\Theta)$.

In the following, we compare a classical evidential pattern mining approaches such as EDMA~\cite{moi16} and U-Apriori~\cite{Chui07} with the output of OpMiner. To do so, we compare these three algorithms in terms of number of extracted patterns and computational time. Figure \ref{frequent pattern} illustrates the number of extracted patterns with regards to the threshold $minsup$. It is evident that the pattern mining approach EDMA finds the highest number of patterns for all fixed $minsup$ comparatively to probabilistic approach approach and OpMiner. In fact, EDMA computes frequent patterns from a set of $28\times 2^5$ items (i.e. sum of the size of all superset of attributes). Therefore, EDMA extracts more patterns than the probabilistic U-Apriori that mines from a set of $28\times 5$ items (i.e. sum of the size of all frames of discernment). OpMiner is has a different approach since an item is a BBA and therefore the number of items is the number of BBAs in the database (i.e., $28\times 11$). In addition, this number is reduced by selecting, at first, only the more committed BBAs. As a result OpMiner is more efficient than the two other approaches since it generates less candidates.
OpMiner not only generates less frequent patterns but more informative ones since it regroups several information in a single item. Even if in our application, all treated BBAs are \textit{simple}\footnote{A BBA is said to be simple if it has at most two focal sets and, if it has two, $\Theta$ is one of those.}, OpMiner works perfectly on \textit{normal} BBAs\footnote{A BBA is said to be normal if $\emptyset$ is not a focal set.}.

	\begin{figure}[!ht]

				       \begin{minipage}[b]{0.45\linewidth}  
						
	\begin{flushleft}					
\begin{tikzpicture}
\begin{axis}[
		scale=0.55,
    enlargelimits=0.1,
		ymode=log,
    ylabel={\# Frequent patterns},
		every axis y label/.style={
    at={(ticklabel* cs:1.08)},
    anchor=south,
},
    symbolic x coords={0.15,0.2,0.3,0.4,0.5},
		xlabel={minsup},
		xmax=0.5,
		xmin=0.15,
legend style={at={(0.6,0.45)},anchor=south west,draw=black,fill=white,legend cell align=left,font=\small},
    every node near coord/.append style={font=\tiny},
   nodes near coords align={vertical},
    ]
\addplot coordinates { (0.5,7) (0.4,30) (0.3,273) (0.2,6864) (0.15, 25013)};
\addplot coordinates { (0.5,18) (0.4,94) (0.3,666) (0.2,10264) (0.15,85234)};
\addplot coordinates { (0.5,20) (0.4,125) (0.3,967) (0.2,16079) (0.15,118079)};
\legend{OpMiner,  U-Apriori~\cite{Chui07},EDMA~\cite{moi16}}

\end{axis}
\end{tikzpicture}

\caption{Number of retrieved frequent patterns from the database.}
\label{frequent pattern}
\end{flushleft}
   \end{minipage}~~~~~~
      \begin{minipage}[b]{0.45\linewidth}
	\begin{flushleft}					
			
		\begin{tikzpicture}
\begin{axis}[
		scale=0.55,
    enlargelimits=0.1,
		ymode=log,
    ylabel={Time (s)},
		every axis y label/.style={
    at={(ticklabel* cs:1.10)},
    anchor=south,
},
    symbolic x coords={0.15,0.2,0.3,0.4,0.5},
		xlabel={minsup},
		xmax=0.5,
		xmin=0.15,
		ymax=10000,
		ymin=0,
			legend style={at={(0.6,0.45)},anchor=south west,draw=black,fill=white,legend cell align=left,font=\small},
    every node near coord/.append style={font=\tiny},
   nodes near coords align={vertical},
    ]
\addplot coordinates {(0.5,0.02) (0.4,0.02) (0.3,0.3) (0.2,32.43) (0.15,530)};
\addplot coordinates {(0.5,0.02) (0.4,0.05) (0.3,0.5) (0.2,43.65) (0.15, 4573)};
\addplot coordinates {(0.5,0.02) (0.4,0.07) (0.3,0.93) (0.2,99.38) (0.15,8546.38)};
\legend{OpMiner,U-Apriori~\cite{Chui07},EDMA~\cite{moi16}}

\end{axis}
\end{tikzpicture}
\caption{Number of retrieved valid association rules from the database.}
\label{association rules}
\end{flushleft}
	\end {minipage}
	\end{figure}

In order to test the quality of the patterns, we oppose the best pattern of EDMA relatively to the first four attributes shown in Table \ref{EDMA's pattern Vs. OpMiner's pattern} to the best one provided by OpMiner.  In fact, it is possible to select from the set of frequent patterns those having items of the four attributes. These patterns show the answer (opinion for OpMiner)  that the majority of the experts have expressed. These patterns are representative of the quality of source $S1$ measures. 
\begin{table}
\centering
\footnotesize
\begin{tabular}{|l|c|c|}
\hline
&EDMA S1 best pattern &OpMiner S1 best pattern\\
\hline
Pattern& \{Q1=Hig or Mod, Q2=Hig or Mod,&\{$ m_1(Mo_1)=1,\begin{cases}m_2(Mo_2)=0.8 \\ m_2(\Theta_2)=0.2\end{cases}$\\
&  Q3=Hig or Mod, Q4=Hig or Mod\} & $m_3(Hig_3)=1, \begin{cases}m_4(Mo_4)=0.8 \\ m_4(\Theta_4)=0.2\end{cases}$\} \\
\hline

\end{tabular}
\caption{EDMA's pattern Vs. OpMiner's pattern}
\label{EDMA's pattern Vs. OpMiner's pattern}
\end{table}
As it is show in Table \ref{EDMA's pattern Vs. OpMiner's pattern}, the construction of both patterns is not the same. EDMA's pattern is constructed from focal elements in contrary of OpMiner that contains BBAs. In addition, the interpretation of both patterns is different. EDMA's pattern shows a hesitation between \textit{high} and \textit{moderate} as an answer trend. Therefore, from this point, making an evaluation of source S1 is not straightforward. OpMiner pattern has a different meaning. It gives for each asked question the most shared opinion (i.e. BBA). It means that, for question 1, 2 and 4 the answer is \textit{moderate} with \textit{high} or \textit{very high} confidence. For question 4, the trend is a high evaluation with a \textit{very high} confidence. As a result, with an overall moderate evaluation of its measure, it is possible to conclude that source S1 is moderately reliable.

\section*{Conclusion}\label{section 4}
In this paper, we introduced a new approach for mining opinion patterns from uncertain database. The uncertainty and the imprecision of the data are modelled with the evidence theory. The extraction is based on new anti-monotonic measures of support derived from the commitment relation.  A mining algorithm OpMiner is then applied to retrieve frequent opinions patterns a from the database. The results on a real-world database shows more informative extracted patterns than literature-based approaches. In future work, we will be interested in refining the inclusion and support measure using the specialization matrix of Smets~\cite{Sme02}. Furthermore, the performance of OpMiner algorithm could be improved by adding specific heuristics such as the decremental pruning\cite{Agg10}.

\subsubsection*{Acknowledgements}
This work is a part of the PEPS project funded by the French national agency for medicines and health products safety (ANSM), and of the SePaDec project funded by Region Bretagne.
\bibliographystyle{splncs}
\bibliography{ref}
\end{document}